\documentclass[letterpaper, 10 pt, conference]{ieeeconf}  

\IEEEoverridecommandlockouts                              

\usepackage{times}
\usepackage{epsfig}
\usepackage{graphicx}
\usepackage{amsmath}
\usepackage{amssymb}
\usepackage{xcolor}
\usepackage[linesnumbered,ruled,vlined]{algorithm2e}
\usepackage{rotating}
\usepackage{pifont}
\usepackage{cuted}
\usepackage{enumitem}
\usepackage{lipsum}
\usepackage[mathscr]{eucal}
\usepackage{array,multirow}
\usepackage{tabularx}
\newcolumntype{s}{>{\hsize=.5\hsize}X}
\definecolor{fgreen}{rgb}{0.13, 0.55, 0.13}
\definecolor{bred}{rgb}{0.55, 0.13, 0.13}
\definecolor{plot1}{HTML}{1f77b4}
\newcommand{\plotblue}[1]{\textcolor{plot1}{#1}}
\definecolor{plot2}{HTML}{ff7f0e}
\newcommand{\plotorange}[1]{\textcolor{plot2}{#1}}
\definecolor{plot3}{HTML}{2ca02c}
\newcommand{\plotgreen}[1]{\textcolor{plot3}{#1}}
\newcommand{\plotred}[1]{\textcolor{red}{#1}}

\DeclareMathOperator*{\argmax}{argmax}

\newcommand{\lturn}[1]{\begin{turn}{90} #1 \end{turn}}

\usepackage{xspace}
\makeatletter
\DeclareRobustCommand\onedot{\futurelet\@let@token\@onedot}
\def\@onedot{\ifx\@let@token.\else.\null\fi\xspace}

\def\etal{\emph{et al}\onedot}
\makeatother
\title{\LARGE \bf
Discwise Active Learning for LiDAR Semantic Segmentation
}

\author{Ozan Unal$^{1,*}$, Dengxin Dai$^{2}$, Ali Tamer Unal$^{3}$, and Luc Van Gool$^{1,4,5}$
\thanks{\hspace{-12px} This work was funded by Toyota Motor Europe via TRACE Zurich.}
\thanks{\hspace{-12px} $^{1}$Computer Vision Lab, ETH Zurich; $^{2}$Huawei Technologies, Zurich Research Center; $^{3}$Department of Industrial Engineering, Bo\u{g}azi\c{c}i University; $^{4}$VISICS, ESAT/PSI, KU Leuven; $^{5}$INSAIT, Sofia}%
\thanks{\hspace{-12px} $^{*}${\tt\small ozan.unal@vision.ee.ethz.ch}}
}

\begin{document}

\maketitle
\thispagestyle{empty}
\pagestyle{empty}

\begin{abstract}
While LiDAR data acquisition is easy, labeling for semantic segmentation remains highly time consuming and must therefore be done selectively. Active learning (AL) provides a solution that can iteratively and intelligently label a dataset while retaining high performance and a low budget. In this work we explore AL for LiDAR semantic segmentation. As a human expert is a component of the pipeline, a practical framework must consider common labeling techniques such as sequential labeling that drastically improve annotation times. We therefore propose a discwise approach (DiAL), where in each iteration, we query the region a single frame covers on global coordinates, labeling all frames simultaneously. We then tackle the two major challenges that emerge with discwise AL. Firstly we devise a new acquisition function that takes 3D point density changes into consideration which arise due to location changes or ego-vehicle motion. Next we solve a mixed-integer linear program that provides a general solution to the selection of multiple frames while taking into consideration the possibilities of disc intersections. Finally we propose a semi-supervised learning approach to utilize all frames within our dataset and improve performance.
\end{abstract}

\section{Introduction}

Dense prediction tasks such as LiDAR semantic segmentation require large amounts of labeled data. While new LiDAR frames are easy to acquire, labeling, especially in 3D, is not only tedious but also highly costly. This severely hinders the scalability and deployability of such tasks in any practical application such as autonomous driving. To retain an economical budget, labeling must therefore be done selectively, as exhaustively labeling all acquired frames is simply not feasible. The question then remains, amongst a pool of acquired LiDAR frames, how do we select which parts should be labeled?

Active learning (AL) provides a framework to intelligently decide what data to label and has proven to be widely successful in comparable 2D tasks where similar economical constraints apply~\cite{vezhnevets2012active, siddiqui2020viewal}. In an AL setting, a model is initially trained on a small amount of labeled data. Based on the model's predictions of the unlabeled data (often on its uncertainty), an acquisition function determines which data to query an external human oracle for a label. The human expert labels the selected data points, which then expand the training set. The model can then be retrained and thus the loop restarts until the predetermined budget is reached. Such a framework that incorporates the human expert into the pipeline often drastically reduces labeling costs while retaining performance. In this work, our goal is to investigate AL for LiDAR semantic segmentation.

Before we move forward, as we aim to include a human labeler into our pipeline, it is important to understand the commonly used techniques for labeling LiDAR sequences in order to ground our AL approach in reality and ultimately provide a deployable solution.
Unlike 2D images, LiDAR scans comprise of 3D points that provide coordinates in the real world. This means sequential point clouds can be projected onto a global coordinate system and concatenated to form a denser representation. This common trick can immensely reduce labeling costs~\cite{iccv2019semantickitti, Unal_2022_CVPR, fong2022nuscenes}. Most outdoor scenes are dominated by static structures (e.g. road, building, vegetation) and static objects (e.g. parked vehicles, bicycles) thus when working on concatenated point clouds they only need to be labeled once to have them labeled on all frames. In fact,  labeling the area of a single frame in global coordinates for concatenated points takes less time than labeling two individual frames based on our user study.

\begin{figure}
    \centering
    \includegraphics[width=\columnwidth]{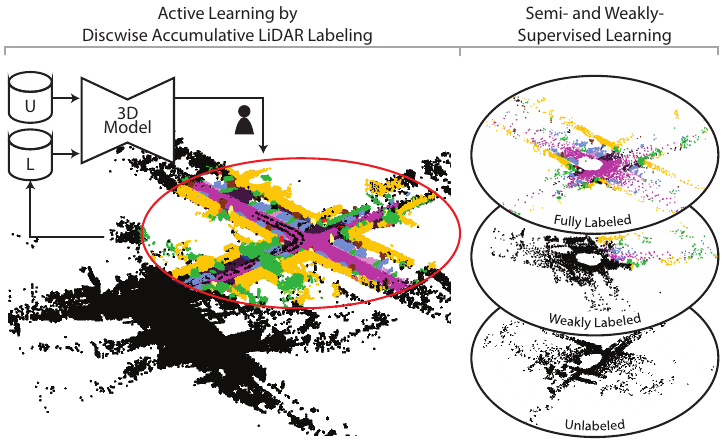}
    \caption{Accumulation of sequential LiDAR point clouds can severely reduce labeling costs, thus an active learning framework must learn to select accumulated regions for labeling to take advantage of this commonly used technique. When discs are queried for labels, the resulting dataset consists of fully labeled, weakly labeled and unlabeled frames.}
    \label{fig:overview}
\vspace{-10px} \end{figure}

To exploit the sequential labeling of concatenated frames within our pipeline and save time labeling, instead of a single frame, we propose constructing the problem of AL for LiDAR semantic segmentation using a disc as the unit data point, i.e. the region a single frame covers on global coordinates. Therefore when a disc is selected to be labeled, the oracle labels all points from any frame that fall within the boundaries of the selected area. An illustration of our proposed discwise labeling can be seen in Fig.~\ref{fig:overview} - left. Exploring discwise AL yields two main challenges which can be identified when compared to the analogous 2D setting.

Firstly, in a common imagewise AL pipeline the pixelwise epistemic uncertainty is aggregated over the image via the sum~\cite{Desai_2022_WACV, saidu2021active} or mean~\cite{ozdemir2021active, gal2017deep} to form the acquisition function. However, unlike 2D images, outdoor LiDAR scenes have vastly varying point densities across a sequence, which can have unintentional consequences when naively aggregating pointwise uncertainty. 
To combat such effects, we propose a simple but intuitive solution where we initially pool information within globally defined voxels via a symmetric function and then aggregate voxelwise uncertainty over discs.

Secondly, following imagewise aggregation, in common 2D AL pipelines the images yielding the highest uncertainty are then queried for labels. However unlike 2D images, discs can intersect with one another in global coordinates and intersecting regions do not provide any additional information. A naive strategy of disallowing intersecting solutions is not guaranteed to yield the optimal disc selection that maximizes uncertainty. Therefore to form a general solution, we construct and solve a mixed-integer linear program that maximizes the information within the total area formed by the selected discs. Our solution allows discs to intersect, if the resulting union would yield a higher uncertainty and collapses to the commonly chosen argmax function when selecting only a single disc.

Having established the major building blocks, we construct a basic yet effective AL pipeline (DiAL) for LiDAR semantic segmentation where a model trains on discwise labeled frames, an acquisition function aggregates uncertainty over discs while respecting point count variability and a selection algorithm selects discs that maximize information.

Finally, we tackle the problem of semi-supervised training via a mean teacher framework~\cite{nips2017meanteacher, Unal_2022_CVPR} to utilize all three types of frames in our actively labeled dataset (see Fig.~\ref{fig:overview}).


\noindent Our contributions can be summarised as follows:
\begin{itemize}
    \setlength\itemsep{0.01em}
    \item We develop a simple active learning pipeline for LiDAR semantic segmentation named DiAL that aligns well with the established techniques for annotating. We query discs on accumulated frames to exploit the effectiveness of sequential data labeling.
    \item We aggregate uncertainty over discs via a two step method to respect the point count variability of outdoor LiDAR data and limit bias introduced via aggregation.
    \item We construct and solve a mixed-integer linear problem that outputs a number of (possibly intersecting) discs, allowing a generalized solution to querying multiple samples each iteration.
    \item With a simple semi-supervised learning approach that utilizes labeled, partially labeled and unlabeled frames, our method performs at $99.8\%$ relative to fully supervised learning on dense labels, with only 50 discs ($\sim0.5\%$ cost of framewise labeling the entire dataset).
\end{itemize}

\section{Related Work}

\noindent \textbf{Active Learning for LiDAR Segmentation:} Active learning provides a framework that leverages existing labeled data points to intelligently decide how to expand the dataset, most commonly via the use of a human oracle~\cite{vezhnevets2012active, siddiqui2020viewal}. The core goal is extract as much performance as possible from a model by labeling as few samples. A commonly used strategy to maximize information yield per labeled data is to sample data points that show high uncertainty. Common metrics include the softmax confidence~\cite{settles2008analysis, wang2016cost}, softmax margin~\cite{roth2006margin, joshi2009multi, Desai_2022_WACV}, softmax entropy~\cite{hwa2004sample, wang2016cost}, or Bayesian approaches that utilize learning by disagreement (BALD)~\cite{kirsch2019batchbald, gal2017deep}.
Recent works have investigated active learning within the context of 3D semantic segmentation~\cite{wu2021redal, hu2022lidal, liu2022less}. Specifically, ReDAL~\cite{wu2021redal} proposes labeling highest information yielding regions within point cloud frames based on softmax entropy, color discontinuity, and structural complexity. However to propose a general AL framework for point cloud segmentation, the sequential nature of outdoor LiDAR scans remain to be considered. LiDAL~\cite{hu2022lidal} takes advantage of this inherent property and utilizes cross-frame predictions as a measure of uncertainty to again, find and label high information providing regions. While both methods shows great performance at low percentage point counts ($1\%-5\%$), we argue that the goal of an AL pipeline should not be to reduce point counts but to reduce labeling times. It is therefore vital to consider the sequential nature of LiDAR frames during the costly labeling stage.

\noindent \textbf{Semi-Supervised LiDAR Semantic Segmentation:} 
LiDAR semantic segmentation remains to be a challenging and computationally
expensive task with current research focused on understanding how to best process the unordered data structure: directly operating on points~\cite{cvpr2017pointnet, arxiv2017pointnet++, wacv2021improving}, projecting it onto 2D~\cite{squeezeseg, squeezesegv2} and the now prevailing strategy, voxelization and 3D sparse convolution~\cite{cvpr2021cylindrical, eccv2020spvnas, cvpr2019minkowski}.
Recently, data-efficient methods have risen in popularity that aim to reduce the labeling cost of the dense prediction task. There are two common strategies: (i) weakly-supervised that provides weak labels for all available frames (e.g. scribble labels with ScribbleKITTI~\cite{Unal_2022_CVPR}), (ii) semi-supervised that provides dense labels for only a subset of frames, while the rest remain unlabeled~\cite{iccv2021guided}. For semi-supervised learning, SemiSup~\cite{iccv2021guided} proposes using a pseudo-label guided point contrastive loss and SSPC~\cite{arxiv2021sspc} utilizes self-training to reduce the gap to their fully supervised baselines.

\section{Active Learning for LiDAR Segmentation}

For LiDAR semantic segmentation, given the vast differences in both time and budgetary requirements for data acquisition and labeling, the latter must be done selectively. An active learning framework provides a solution to intelligently decide what data to label next. In the following sections, we explore the individual steps that typically form the loop of an active learning framework:

\noindent \textbf{\ref{sec:pointwise_uncertainty}} A model, trained on the labeled dataset, is used to determine pointwise uncertainty for the unlabeled data.

\noindent \textbf{\ref{sec:discwise_aggregation}} The uncertainty is aggregated over the chosen unit of data, which in our case is the accumulated region (disc).

\noindent \textbf{\ref{sec:disc_selection}} The data points yielding the highest uncertainty are queried for labels and thus expand the labeled dataset.

\subsection{Pointwise Epistemic Uncertainty Estimation}  \label{sec:pointwise_uncertainty}

The goal in LiDAR semantic segmentation is to discover the dependency of the pointwise distribution over the labels $y \in Y$ on an input variable $x \in X$ via the model weights $w$. Formally, reducing this to a classification task where $y$ can be of class $c \in C$, we define the conditional probability as:

\begin{equation}
    p(y=c|x,X,Y) = \int p(y=c|x,\omega) p(\omega|X,Y) d\omega
\end{equation}

As the posterior distribution $p(w|X,Y)$ is intractable, we approximate it with a variational distribution $q(\omega)$ and aim to minimize the Kullback-Leibler divergence between the two distributions KL$(q(\omega) | p(\omega | X, Y))$. Following Gal.~\etal~\cite{gal2017deep}, we further approximate the variational distribution via Monte Carlo integration by employing variational inference, i.e. by applying stochastic forward passes (i.e. a model is trained with dropout and dropout is performed during inference):
\begin{equation}
\begin{split}
    p(y=c|x,X,Y) & \approx \int p(y=c|x,\omega) q(\omega) d\omega \\
     & \approx \frac{1}{N}\sum_{n=1:N}p(y=c|x,\hat\omega_n)
    \label{eq:mc_integration}
\end{split}
    \end{equation}

We measure the pointwise epistemic uncertainty within our training set $X$ as the mutual information between the model parameters and the model posterior:
\begin{equation}
    \mathbb{I}[y,\omega|x,X] = \mathbb{H}[y|x,X]-\mathbb{E}_{p(\omega|X)} \left[ \mathbb{H}[y|x,\omega]  \right]
    \label{eq:mutual_information}
\end{equation}
with ${\mathbb{H}[y|x,X] = -\sum_c p(y=c|x,X) \log p(y=c|x,X)}$ defining the pointwise entropy. Inserting Eq.~\ref{eq:mc_integration} into Eq.~\ref{eq:mutual_information}:
\begin{equation}
\begin{split}
        \mathbb{I}[y,\omega|x,X] \approx& -\sum_c \left ( \frac{1}{N} \sum_n \hat{p_c} \right ) \log \left ( \frac{1}{N} \sum_n \hat{p_c} \right ) \\
        &+ \frac{1}{N}\sum_c \sum_n \hat{p_c} \log \hat{p_c}
\end{split}
\end{equation}
with $\hat{p}$ denoting the softmax output of a stochastic forward pass. For the derivation, please refer to \cite{gal2016dropout}. 




\subsection{Discwise Uncertainty Aggregation}  \label{sec:discwise_aggregation}
Having established a metric to capture pointwise uncertainty, the next step in an active learning pipeline is to aggregate this information over a chosen \textit{unit of data}. This aggregate will then be used to guide the ensuing data unit selection step. The choice of the data unit is therefore not trivial as it will determine what will be presented to the human annotator. To make a practical choice, it is important to first understand the labeling process of LiDAR point clouds for semantic segmentation to ground our choice in reality and take advantage of commonly utilized techniques.

LiDAR labeling is notoriously difficult and requires a high level of expertise from the annotator. While scenes are represented in 3D, labeling is done on 2D projections onto a monitor screen. As the projection is view dependent, the labeling process requires constant navigation and readjustments within the 3D space. Labeling a 2D area on a screen labels every point that falls within that selected area, commonly causing unintentional regions to be labeled, thus requiring regular corrections.

While working with 3D point clouds does come with obvious disadvantages for labeling, there is one strength that drastically speeds up the process. Given sequential data (as commonly acquired in outdoor LiDAR applications), individual frames can be projected onto a global coordinate system via odometry poses to form a concatenated point cloud sequence. This allows each global area to be labeled once, in order to have it labeled on every frame. Given that most outdoor scenes are dominated by static things (e.g. parked vehicles) and static stuff (e.g. road, building), this technique can save thousands of hours for large-scale datasets. To quantify the impact of this method, we utilize in-house annotators to label multiple frames as well as their corresponding discs. We define a disc as the region a single frame would cover on global coordinates (see Fig.~\ref{fig:overview} - left for an illustration of a labeled disc).  We observe that with an overhead of $10\%-80\%$\footnote{For SemanticKITTI~\cite{iccv2019semantickitti}, we observe that a disc takes approximately 3.5 hours to label due to their fixed range, while the corresponding frame can vary significantly based on content and environment (which is in line with the reported labeling times of accumulated tiles~\cite{iccv2019semantickitti}).}, one can not only fully label a single frame, but also weakly label several other frames within the same sequence by utilizing concatenated point cloud sequences.  Unlike previous work that focus on reducing the total number of points labeled~\cite{wu2021redal, hu2022lidal}, our goal in this work is to minimize labeling times. Therefore following this observation, to take advantage of sequential labeling, we propose using discs as units of data within our AL pipeline.

To aggregate pointwise uncertainty over a disc, we first consider 2D analogous aggregation methods such as the sum~\cite{Desai_2022_WACV, saidu2021active} or the mean~\cite{ozdemir2021active, gal2017deep}. However we note that unlike 2D images that have constant pixel counts, LiDAR discs can have vastly different point counts which can bias such methods. In specific, the \textit{mean} fails to consider the variation in discwise point counts due to environmental changes (e.g. highway scenes have an overall low number of points compared to in-city 4-way junctions), and the \textit{sum} fails to consider local density changes due to the variation in ego vehicle velocity (e.g a slow moving vehicle would accumulate more frames and thus more points within the same global environment compared to a fast moving car).

To combat these effects, we propose an intuitive two step aggregation method.

Firstly, to combat the local density changes, we aggregate mutual information (MI) within local neighborhoods via a symmetric function, i.e. we pool within globally defined voxels. Formally, we define the aggregated voxelwise MI as:
\begin{equation}
     \mathbb{I}_j = f_{x_i \in V_j} \mathbb{I}[y_i,w|x_i,X]
\end{equation}
with the set $V_j$ containing all points $x_i$ within the voxel centered around $v_j$ of length $l_{voxel}$, i.e. $V_j = \{x_i \, | \, ||x_i - v_j|| \leq l_{voxel} / 2\}$ and $f$  chosen as min or max~\cite{cvpr2017pointnet, arxiv2017pointnet++}.

Having voxelwise uncertainty, we can then aggregate the mutual information within discs to form our acquisition metric to guide our decision on what region to label next. To combat discwise point count variations we refer back to the sum. Formally, we define the discwise uncertainty $\alpha_i$ as:
\begin{equation}
    \alpha_i = \textstyle{\sum_j} \mathbf{z}_{j,i} \mathbb{I}_j
\end{equation}
with $\mathbf{z}_{j,i}$ defining a binary mask of voxels within the disc of radius $R$ centered around $x_i$:
\begin{equation} \label{eq:z_ji}
    z_{j,i} = 
    \begin{cases}
    1\textrm{,  if  } || v_j - x_i || - R \leq 0 \\
    0\textrm{,  otherwise.}
    \end{cases}
\end{equation}

With the two step aggregation, we aim to maximize the pooled information we obtain from globally defined local neighborhoods based on the conditions set by function $f$.

It should be noted that, we consider a disc as the unit of data to remain comparable to current state-of-the-art framewise active learning strategies. However our method provides a general solution to the selection of any region shape within the registered point cloud sequence, as Eq.~\ref{eq:z_ji} can be freely designed based on annotator preferences.

\begin{figure} \vspace{5px}
    \centering
    \includegraphics[width=\columnwidth]{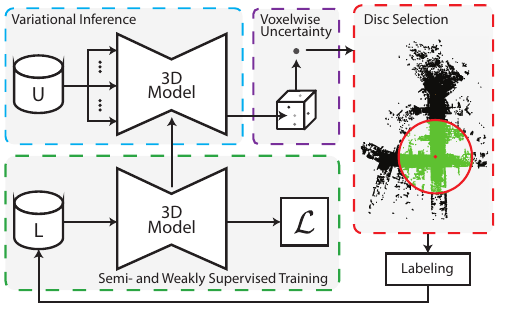}
    \caption{Illustration of the active learning pipeline. We first estimate pointwise epistemic uncertainty, aggregate it on globally defined voxels and select discs that maximize the total information. The points within each disc is then labeled by an annotator.}
    \label{fig:pipeline}
\vspace{-10px} \end{figure}

\subsection{Disc Selection for Labeling} \label{sec:disc_selection}

Given discwise uncertainty, we can now query the disc $\hat{i}$ that would yield the highest amount of new information:
\begin{equation} \label{eq:single_disc_selection}
    \hat{i} = \argmax \alpha_i.
\end{equation}

Selecting a single disc to label in each iteration of the active learning algorithm comes with additional fix costs that are associated with not only the initiation of the data labeling pipeline (e.g. set up time for the annotator) but also with the training of the model. A common solution to this issue is to query multiple data samples in each iteration~\cite{papadopoulos2016we, vezhnevets2012active, wu2021redal}.

While the selection of a single disc can be done via Eq.~\ref{eq:single_disc_selection}, the selection of multiple discs in each iteration cannot be trivially handled. Unlike with 2D images, discs can intersect in global coordinates and intersecting regions do not provide any additional information to the system. Yet, naively prohibiting intersections within the selection of discs would not guarantee the highest information yield.

To this end, we propose a general solution by constructing a mixed-integer linear program that maximizes the information within the union of the areas formed by the discs. To select discs centered around possible points $x_i$ of radius $R$, we solve the following problem:
\begin{equation} \label{eq:mixed-integer}
\begin{aligned}
    \max \quad & \textstyle{\sum_j} \mathbf{v}_j \mathbb{I}_j \\
    \textrm{s.t.} \quad & \textstyle{\sum_i} \mathbf{x}_i = N  \\
    & \mathbf{v}_j \leq \textstyle{\sum_i} z_{j,i} \mathbf{x}_i, \ \forall j \\
    & 0 \leq \mathbf{v}_j \leq 1, \ \forall j  \\
    & \mathbf{x}_i \in \{0,1\}, \ \forall i 
\end{aligned}
\end{equation}
with $\mathbf{x}_i$ and $\mathbf{v}_j$ denoting binary and real variables that determine if a certain center point or voxel has been selected.

Here, the first constraint fixes the total number of selected discs to a predetermined value of $N$ for each iteration, and the second constraint ensures that we select the union of the disc masks $z_{j,i}$ (see Eq.~\ref{eq:z_ji}) for selected disc centers $x_i=1$. Although $\mathbf{v}_j$'s are defined as $\{\mathbf{v}_j \in \mathbb{R} \, |\, 0 \leq \mathbf{v}_j \leq 1 \}$, the $\mathbf{x}_i$ variables being binary guarantees that $\mathbf{v}_j$'s would assume binary values in any optimal solution.

In ensuing steps of the AL pipeline, we set an additional constraint to fix the previously selected discs $\hat{i}$ ($x_{\hat{i}} = 1 \ \forall \ \hat{i}$) and adjust the first constraint accordingly ($\textstyle{\sum_i} \mathbf{x}_i = tN$ for time step $t$). Our overall proposed AL framework for LiDAR semantic segmentation is illustrated in Fig.~\ref{fig:pipeline}.

\begin{figure} \vspace{5px}
    \centering
    \includegraphics[width=\columnwidth]{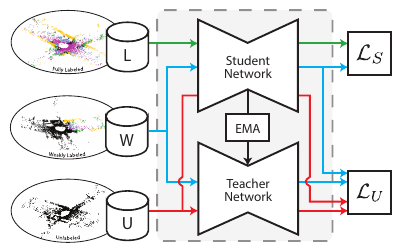}
    \caption{Mean teacher framework for semi-supervised LiDAR semantic segmentation to take advantage of fully labeled (L), weakly labeled (W) and unlabeled (U) frames within the overall dataset. Colored paths show the distinct paths for the types of frames.}
    \label{fig:semi_supervised}
\vspace{-10px} \end{figure}

\section{Semi-Supervised LiDAR Segmentation}

By labeling discs, i.e. during any point in our active learning pipeline, our overall dataset consists of three types of frames: (i) fully labeled frames that span a complete disc region, (ii) weakly labeled that partially intersect with the labeled disc in global coordinates and (iii) unlabeled frames. When naively training with a supervised loss, we only make use of labeled points. To better utilize all information within our dataset, we construct a semi-supervised training approach that can utilize all forms of information to improve the quality of the final dense predictions of our model. 

Points of all frame types can be divided into two subsets: labeled, and unlabeled. For labeled points we apply a supervised loss $H$ that is typically chosen as cross entropy:
\begin{equation} \label{eq:supervised_loss}
    \mathcal{L}_S = H(\hat{y}, y)
\end{equation}
given the predicted class distribution $\hat{y}$, and ground truth $y$. 

To extend the supervision to unlabeled points we utilize a mean teacher framework (MT)~\cite{nips2017meanteacher}. The MT framework consists of a student that is typically trained using backpropagation (see Eq.~\ref{eq:supervised_loss}), and a teacher network who's weights ($\theta^\textrm{EMA}$) are computed as the exponential moving average (EMA) of the student's.
\begin{equation} \label{eq:ema}
    \theta^\textrm{EMA}_t = \beta \theta^\textrm{EMA}_{t-1} + (1-\beta) \theta_t
\end{equation}

To make use of the more robust representation capabilities of the moving average weights, we apply a consistency loss between the student's predictions and the teacher's via~\cite{Unal_2022_CVPR}:
\begin{equation} \label{eq:unsupervised_loss}
    \mathcal{L}_U = \textrm{KL}(p(\hat{y}_i), p(\hat{y}^\textrm{EMA}_i))
\end{equation}
with $\hat{y}^\textrm{EMA}$ denoting the teacher's predicted class distribution. To increase the variation between the student and teacher's outputs and therefore more effectively utilize the unsupervised loss, we apply perturbations to the student's input in the form of rotation, Gaussian jitter, translation and scaling.

\noindent Formally, we define the total loss as:
\begin{equation}
    \mathcal{L} = \mathcal{L}_S + \mathcal{L}_U
\end{equation}
Our proposed mean-teacher based semi-supervised learning pipeline is illustrated in Fig.~\ref{fig:semi_supervised}.

\section{Experiments}

\noindent \textbf{Implementation Details:} In our experiments we use the popular SPVCNN~\cite{eccv2020spvnas} as a baseline model which already includes dropout layers. We use GUROBI~\cite{gurobi} to solve the optimization problem. We do 10 stochastic forward passes for each sample to compute the mutual information. We set a radius of $50m$ for discs following precedent on LiDAR frame processing~\cite{cvpr2021cylindrical, eccv2020spvnas} and a voxel size of $0.5m$ for the initial local aggregation. The disc size allows us to remain comparable to framewise methods. Furthermore, querying large regions allows us to better capture the underlying long tailed distribution of driving scenes.

We select $N=5$ discs on each step, starting with an initial state of 1 arbitrarily chosen disc (sequence 0, disc 100). We set $\beta = 0.99$ following Unal~\etal~\cite{Unal_2022_CVPR}. We set $f$ as the min-operation unless stated otherwise.
Furthermore, in this work we limit the possible disc centers $x_i$ to the ego vehicle path to reduce the search space. It should be noted that while we impose this constraint for our experiments, our AL pipeline along with the mixed-integer linear program in Eq.~\ref{eq:mixed-integer} forms a general solution, allowing a free choice $x_i$ and even $z_{j,i}$.

\noindent \textbf{Dataset:} We test and extensively ablate our method on SemanticKITTI~\cite{iccv2019semantickitti} which is the most popular large-scale autonomous driving dataset for LiDAR semantic segmentation. It consists of 11 sequences with publicly available labels, among which sequence 8 is reserved for validation. The training sequences consist of 19130 frames. Evaluation is carried via the mIoU metric over 19 classes on the \textit{val}-set. We also showcase that our method works on two further datasets ScribbleKITTI~\cite{Unal_2022_CVPR} and nuScenes~\cite{fong2022nuscenes} and with other baseline models (MinkowskiNet~\cite{cvpr2019minkowski}), with similar performance gains over random disc selection.




\begin{table}[t]
    \centering
    \includegraphics[width=\columnwidth]{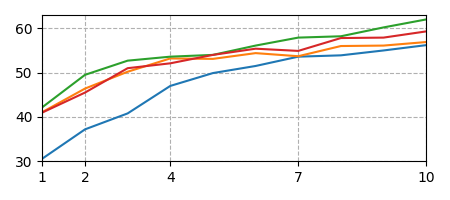}
    \begin{tabular}{|l|cccccc|}
    \hline
    Steps & 0 & 1 & 2 & 4 & 7 & 10 \\
    \hline
    \plotblue{CONF} & 25.7 & 30.6 & 37.2 & 47.0 & 53.6 & 56.2 \\
    \plotorange{MAR} & 25.7 & 41.1 & 46.4 & 53.2 & 53.7 & 56.9\\
    \plotred{ENT} & 25.7 & 41.0 & 45.5 & 52.1 & 54.9 & 59.3\\
    \plotgreen{MI} & 25.7 & \textbf{42.2} & \textbf{49.5} & \textbf{53.6} & \textbf{57.9} & \textbf{62.0}\\
    \hline
    \end{tabular}
    \caption{Comparison of different uncertainty metrics: (i) the softmax confidence (CONF), (ii) the softmax margin (MAR), and (iii) the softmax entropy (ENT) to mutual information (MI).}
    \label{tab:uncertainty}
\vspace{-15px} \end{table}

\begin{table}[t]
    \centering
    \tabcolsep=0.2cm
    \begin{tabular}{|cl|ccccc|}
    \hline
    & Exp. Time &  $1\%$ & $2\%$ & $3\%$ & $4\%$ & $5\%$ \\
    & Labeled &  ($1\%$) & $2\%$ & $3\%$ & $4\%$ & $5\%$ \\
    \hline
    \multirow{7}{*}{\rotatebox[origin=c]{90}{Framewise}}
    & RAND$_\textrm{fr}$ & 48.8 & 52.1 & 53.6 & 55.6 & 57.2 \\
    & RAND$_\textrm{re}$ & 48.8 & 51.7 & 55.0 & 56.1 & 58.2 \\
    & SEGENT & 48.8 & 49.8 & 48.3 & 49.1 & 48.2 \\ 
    & CSET & 48.8 & 53.1 & 52.9 & 53.2 & 52.6  \\ 
    & ReDAL~\cite{wu2021redal} & 48.8 & 51.3 & 54.0 & 58.6 & 58.1 \\
    & LiDAL~\cite{hu2022lidal} & 48.8 & 57.1 & 58.7 & 59.3 & 59.5 \\
    & DiAL (Ours) & 44.4* & 53.6 & 55.9 & 56.3 & 58.1  \\ 
    \hline
    \hline
    & Exp. Time &  $1\%$ & $2\%$ & $3\%$ & $4\%$ & $5\%$ \\
    & Labeled & $60\%$ & $90\%$ & $99\%$ & $99\%$ & $99\%$\\
    \hline
    \multicolumn{2}{|c|}{DiAL (Discwise Ours)} & \textbf{61.4} & \textbf{63.8} & \textbf{63.8} & \textbf{63.8} & \textbf{63.8} \\
    \hline
    \end{tabular}
    \caption{Comparison of AL methods under fixed budget constraints. *~indicates a different initial state.} 
    \label{tab:disc_vs_current}
\vspace{-20px} \end{table}

\begin{table}[t] \vspace{5px}
    \centering
    \tabcolsep=0.12cm
    \begin{tabular}{|l|cccccc|}
    \hline
    Steps & 0 & 1 & 2 & 4 & 7 & 10 \\
    \hline
    Framewise &  1.1 & 19.2 & 22.6 & 26.1 & 32.6 & 40.0 \\
    Discwise & \textbf{25.7} & \textbf{42.2} & \textbf{49.5} & \textbf{53.6} & \textbf{57.9} & \textbf{62.0}\\
    $\Delta$ & \textbf{\color{fgreen}+24.6} & \textbf{\color{fgreen}+23.0} & \textbf{\color{fgreen}+26.9} & \textbf{\color{fgreen}+27.5} & 
    \textbf{\color{fgreen}+25.3} & 
    \textbf{\color{fgreen}+22.0} \\
    \hline
    \end{tabular}
    \caption{Comparison of framewise and discwise labeling under realistic low budget constraints.} 
\label{tab:disc_vs_frame}
\vspace{-20px} \end{table}

\begin{table}[t]
    \centering
    \tabcolsep=0.3cm
    \begin{tabular}{|c|cccc|}
    \hline
    Labeled & 1\% & 3\% & 6\% & 12\%\\
    \hline
    ReDAL [M39] & 25.7 & 37.7 & 42.1 & 47.6 \\
    LiDAL [M12] & 25.7 & 40.2 & 45.8 & 49.9 \\
    DiAL (Ours) & 25.7 & \textbf{42.2} & \textbf{49.5} & \textbf{53.6} \\
    \hline
    \end{tabular}
    \caption{Comparison of AL with sequential labeling.}
    \label{tab:fairer_comparison}
\vspace{-20px}\end{table}

\begin{table}[t] \vspace{5px}
    \centering
    \includegraphics[width=.9\columnwidth]{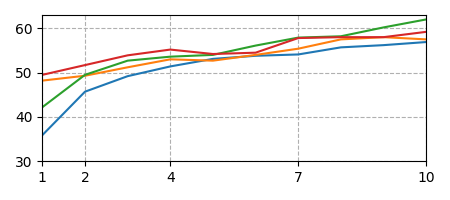}
    \tabcolsep=0.18cm
    \begin{tabular}{|cc|cccccc|}
    \hline
    $f$ & D & 0 & 1 & 2 & 4 & 7 & 10 \\
    \hline
    - & \plotblue{Mean} & 25.7 & 35.9 & 45.7 & 51.4 & 54.1 & 56.9 \\
    - & \plotorange{Sum} & 25.7 & 48.2 & 49.3 & 53.0 & 55.4 & 57.5 \\
    \plotred{Max} & \plotred{Sum} & 25.7 & \textbf{49.5} & \textbf{51.7} & \textbf{55.2} & 57.8 & 59.2 \\
    \plotgreen{Min} & \plotgreen{Sum} & 25.7 & 42.2 & 49.5 & 53.6 & \textbf{57.9} & \textbf{62.0} \\
    \hline
    \end{tabular}
    \caption{Comparison of disc aggregation strategies. $f$ denotes the symmetric function and D denotes the discwise aggregation.}
    \label{tab:disc_aggregation}
\vspace{-10px}\end{table}

\subsection{Results and Ablation Studies} \label{sec:ablation}
In this section, we conduct extensive experiments and ablation studies to isolate the effect of each of our proposed components in DiAL, and compare them to both baseline methods and current approaches in the literature in order to show their necessity and effectiveness for discwise AL.

\noindent \textbf{Uncertainty Metric:} We compare the mutual information to other metrics commonly used for uncertainty estimation for classification tasks, including (i) the softmax confidence (CONF), (ii) the softmax margin (MAR), and (iii) the softmax entropy (ENT). More detailed explanation on all baseline approaches can be found in Hu~\etal~\cite{hu2022lidal}. As seen in Tab.~\ref{tab:uncertainty}, the mutual information provides strictly better performance within our proposed active learning pipeline for LiDAR semantic segmentation, and thus we retain this metric for the remainder of the experiments.

\noindent \textbf{Disc vs. Frame:} We compare our proposed strategy of discwise active learning to methods that do not take sequential labeling into consideration, including random selection of frames (RAND$_\textrm{fr}$), random selection of framewise regions (RAND$_\textrm{re}$), segment-entropy (SEGENT), core-set selection (CSET), ReDAL~\cite{wu2021redal} and LiDAL~\cite{hu2022lidal}. A detailed explanation on baseline approaches can be found in Hu~\etal~\cite{hu2022lidal}.~\footnote{To compare to the aforementioned framewise methods under a fixed labeling budget, we need to first determine the equivalent number of discs to label, as, to emphasize once again, matching the percentage of labeling points would not match the labeling times. Assuming point percentage is directly correlated with frame percentage, to label $1\%$ of points, we determine the appropriate number frames to label as $1\%$ of the total frame count. To compensate for the time difference between discwise and framewise labeling, we then apply our active learning strategy with discwise selection until a conservative $(1\%) F / 2$ discs are queried, with $F$ denoting the total frame count in the \textit{train}-set. For the initial state, we randomly sample discs while minimizing intersections. We therefore ensure that each method is compared under the same labeling budget.}

Firstly in Tab.~\ref{tab:disc_vs_current} - top, we compare our approach applied with framewise active learning to existing approaches. As seen, our baseline framewise active learning framework performs on par with existing methods such as ReDAL~\cite{wu2021redal}.

Second, we showcase the effectiveness of sequential labeling in an AL setting in Tab.~\ref{tab:disc_vs_current} - bottom, where we observe significantly better performance compared to our framewise AL baseline under a fixed labeling budget.


Of course these results are completely expected when we consider the percentage of points labeled. Here, with the aid of Tab.~\ref{tab:disc_vs_current} we would like draw the readers attention to how misleading the percentage of labeled points might be when comparing it to labeling times of LiDAR point clouds. While the reported percentages may seem sparse, by the time it takes to label $1\%$ of points on individual frames (191 frames), around $60\%$ of the total point count can be labeled (95 discs)\footnote{The point count of each disc can correspond to up to 100 times the point count of the corresponding frame. We note that this ratio holds for most datasets as ego-vehicles often move at low velocities during data acquisition. Even at higher velocities, we argue that this example provides sufficient grounds that emphasizes the benefits of sequential labeling.}. These findings once again emphasize the need to ground LiDAR based active learning research in commonly used techniques for labeling to ensure practicality.
To yet again showcase the advantages of discs over frames in low budget scenarios, we also provide an additional comparison in Tab.~\ref{tab:disc_vs_frame} where we see an overwhelming advantage of sequential labeling within AL. 

\begin{table}[t]
    \centering
    \includegraphics[width=.9\columnwidth]{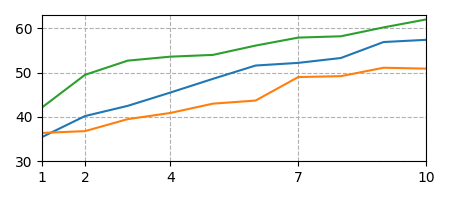}
    \begin{tabular}{|l|cccccc|}
    \hline
    Steps & 0 & 1 & 2 & 4 & 7 & 10 \\
    \hline
    \plotblue{RAND} & 25.7 & 35.5 & 40.2 & 45.5 & 52.2 & 57.4 \\
    \plotorange{HPCS} & 25.7 & 36.4 & 36.8 & 40.9 & 49.0 & 50.9\\
    \plotgreen{Ours} & 25.7 & \textbf{42.2} & \textbf{49.5} & \textbf{53.6} & \textbf{57.9} & \textbf{62.0} \\
    \hline
    \end{tabular}
    \caption{Comparison of disc selection strategies (i) random (RAND), (ii) highest point count selection (HPCS) and (iii) ours.}
    \label{tab:disc_selection}
\vspace{-20px} \end{table}

\noindent \textbf{Adapting SOTA to sequential AL:} We strongly argue that all AL methods \textit{should} utilize sequential labeling but not all of them are naturally extendable to incorporate sequential labeling. Specifically, ReDAL and LiDAR rely on memory heavy region extraction methods limiting them from being used on accumulated point clouds. Still to exploit sequential labeling, we construct a new pipeline where we use ReDAL and LiDAL to select a region to label, fit a convex hull onto this region and label all points within the hull in the accumulated point cloud sequence. This allows DiAL, which is developed directly with sequential labeling at its core, to better compare to baseline approaches under sequential labeling. In Tab.~\ref{tab:fairer_comparison} we observe that our discwise AL strategy outperforms both work. This is because both LiDAL and ReDAL often pick distant sparse regions that show a higher uncertainty in the current frame, however when sequential labeling, also contain a large number of high certainty points from other frames. Specifically, compared to DiAL, LiDAL and ReDAL tend to sample more head class heavy regions (e.g. regions of mainly road, building points) as they can show low uncertainty in certain frames at long ranges. However in the sequential setting, this yields a non-optimal selection and thus a reduced performance for tail classes. While head classes perform similarly across different methods (the IoU for road, building, vegetation are within $3\%$), we observe large gains in tail classes such as bicycle (up to $+22.5\%$), other vehicle (up to $+17.7\%$, and person (up to $8.3\%$) when directly optimizing for sequential labeling.

\begin{table*}[t] \vspace{5px}
    \tabcolsep=0.11cm
    \resizebox{\textwidth}{!}{
    \begin{tabular}{|l|c|ccccccccccccccccccc|}
        \hline
        Method
        & mIoU
        &\lturn{car}
        &\lturn{bicycle}
        &\lturn{motorcycle}
        &\lturn{truck} 
        &\lturn{other vehicle } 
        &\lturn{person}
        &\lturn{bicyclist} 
        &\lturn{motorcyclist} 
        &\lturn{road}
        &\lturn{parking}  
        &\lturn{sidewalk} 
        &\lturn{other ground} 
        &\lturn{building} 
        &\lturn{fence}
        &\lturn{vegetation} 
        &\lturn{trunk}
        &\lturn{terrain} 
        &\lturn{pole}
        &\lturn{traffic sign} \\
        [0.5ex]
        \hline
        SPVCNN (Dense labels)~\cite{eccv2020spvnas} & 63.8 & 97.1 & 35.2 & 64.6 & 72.7 & 64.3 & 69.7 & 82.5 & 0.2 & 93.5 & 50.8 & 81.0 & 0.3 & 91.1 & 63.5 & 89.2 & 66.1 & 77.2 & 64.1 & 49.4 \\
        \hline
        RAND & 57.4 & 95.1 & 8.2 & 56.1 & 69.5 & 31.1 & 62.7 & 80.8 & 0.0 & 91.8 & 34.2 & 76.8 & 0.1 & 91.4 & 61.7 & 88.6 & 55.5 & 76.5 & 62.9 & 48.3 \\
        HPCS & 50.9 & 93.9 & 5.3 & 31.7 & 29.6 & 28.3 & 53.9 & 51.0 & 0.0 & 92.3 & 40.0 & 78.5 & 0.0 & 90.2 & 55.2 & 87.2 & 57.5 & 76.1 & 60.6 & 36.0 \\
        DiAL (Supervised) & 62.0 & 96.1 & 30.1 & 62.7 & 69.7 & 54.4 & 68.5 & 85.5 & 0.0 & 89.7 & 45.7 & 76.8 & 0.1 & 91.0 & 61.8 & 88.3 & 66.5 & 75.3 & 64.8 & 50.7 \\
        $\Delta$ (RAND) &  \textbf{\color{fgreen}+4.6} & \textbf{\color{fgreen}+1.0} &
        \textbf{\color{fgreen}+21.9} &
        \textbf{\color{fgreen}+6.6} &
        \textbf{\color{fgreen}+0.2} &
        \textbf{\color{fgreen}+23.3} &
        \textbf{\color{fgreen}+5.8} &
        \textbf{\color{fgreen}+4.7} &
        \textbf{\color{gray}0.0} &
        \textbf{\color{bred}-2.1} &
        \textbf{\color{fgreen}+11.5} &
        \textbf{\color{gray}0.0} &
        \textbf{\color{gray}0.0} &
        \textbf{\color{bred}-0.4} &
        \textbf{\color{fgreen}+0.1} &
        \textbf{\color{bred}-0.3} &
        \textbf{\color{fgreen}+11.0} &
        \textbf{\color{bred}-1.2} &
        \textbf{\color{fgreen}+1.9} &
        \textbf{\color{fgreen}+2.4} \\
        \hline
        DiAL (Semi-supervised) & 63.7 & 96.0 & 42.0 & 63.7 & 82.5 & 46.1 & 74.6 & 86.3 & 0.0 & 93.8 & 44.0 & 79.6 & 13.0 & 89.9 & 56.8 & 87.1 & 66.0 & 71.7 & 65.2 & 51.6 \\
        
        $\Delta$ (Supervised) & \textbf{\color{fgreen}+1.7} & \textbf{\color{bred}-0.1} &
        \textbf{\color{fgreen}+12.1} &
        \textbf{\color{fgreen}+1.0} &
        \textbf{\color{fgreen}+12.8} &
        \textbf{\color{bred}-8.3} &
        \textbf{\color{fgreen}+6.1} &
        \textbf{\color{fgreen}+0.8} &
        \textbf{\color{gray}0.0} &
        \textbf{\color{fgreen}+4.1} &
        \textbf{\color{bred}-1.7} &
        \textbf{\color{fgreen}+2.8} &
        \textbf{\color{fgreen}+12.9} &
        \textbf{\color{bred}-1.1} &
        \textbf{\color{bred}-5.0} &
        \textbf{\color{bred}-1.2} &
        \textbf{\color{fgreen}+0.5} &
        \textbf{\color{bred}-3.6} &
        \textbf{\color{fgreen}+0.4} &
        \textbf{\color{fgreen}+0.9} \\
        \hline
    \end{tabular}
    }
    \caption{Comparison of the current baselines at step 10 (51 discs) to our proposed active learning framework with and without semi-supervised training. Alongside the mIoU, we provide classwise IoU metrics to show the advantages of active learning.
    \label{tab:results}}
\vspace{-20px}\end{table*}

\begin{figure*}
    \centering
    \includegraphics[width=\linewidth]{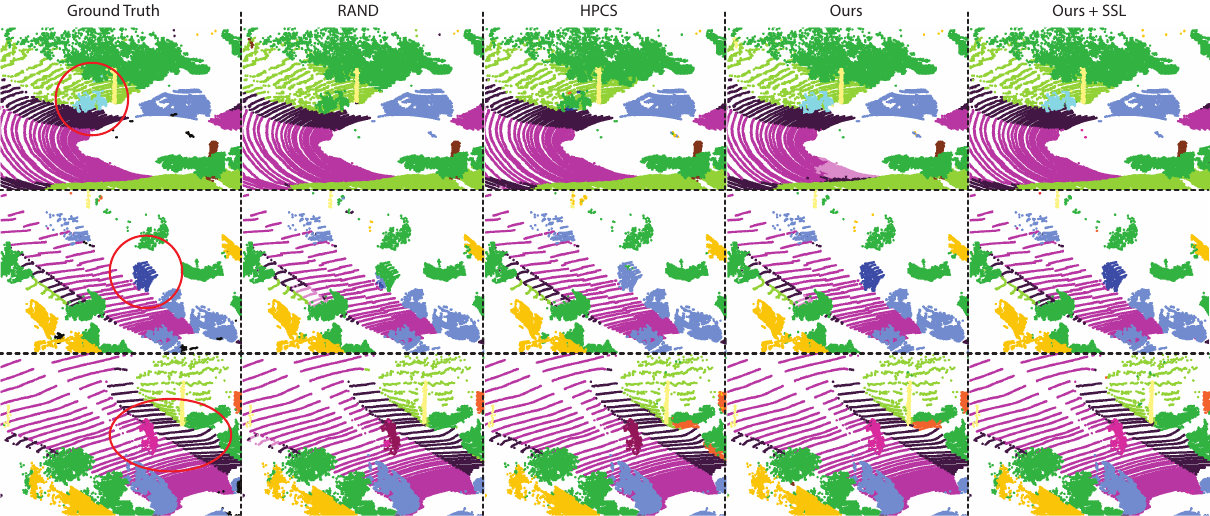}
    \caption{Example results from the SemanticKITTI validation set comparing current baseline approaches to our proposed discwise active learning pipeline trained in a supervised and semi-supervised manner (+ SSL).}
    \label{fig:results}
\end{figure*}

\begin{table*}[t]
\begin{minipage}{.29\textwidth}
    \tabcolsep=0.16cm
    \centering
    \begin{tabular}{|l|ccccc|}
    \hline
    Steps & 1 & 2 & 4 & 7 & 10 \\
    \hline
    \plotblue{RAND} & 35.5 & 37.0 & 44.0 & 48.8 & 53.8 \\
    \plotorange{Ours} & 40.8 & 49.5 & 51.9 & 56.5 & 59.0 \\
    $\Delta$ & \textbf{\color{fgreen}+5.3} & \textbf{\color{fgreen}+12.5} & \textbf{\color{fgreen}+7.9} & \textbf{\color{fgreen}+7.7} & \textbf{\color{fgreen}+5.2} \\
    \hline
    \end{tabular}
    \caption{MinkNet~\cite{cvpr2019minkowski} on SemanticKITTI~\cite{iccv2019semantickitti}.}
    \label{tab:minkowski}
\end{minipage}
\hfill
\begin{minipage}{.29\textwidth}
    \tabcolsep=0.16cm
    \centering
    \begin{tabular}{|l|ccccc|}
    \hline
    Steps & 1 & 2 & 4 & 7 & 10 \\
    \hline
    \plotblue{RAND} & 33.6 & 36.6 & 44.2 & 50.6 & 52.8 \\
    \plotorange{Ours} & 37.7 & 39.4 & 47.3 & 52.4 & 54.4 \\
    $\Delta$ & \textbf{\color{fgreen}+4.1} & \textbf{\color{fgreen}+2.8} & \textbf{\color{fgreen}+3.1} & \textbf{\color{fgreen}+1.8} & \textbf{\color{fgreen}+1.6} \\
    \hline
    \end{tabular}
    \caption{SPVCNN~\cite{eccv2020spvnas} on ScribbleKITTI~\cite{Unal_2022_CVPR}.}
    \label{tab:scribblekitti}
\end{minipage}
\hfill
\begin{minipage}{.29\textwidth}
    \tabcolsep=0.16cm
    \centering
    \begin{tabular}{|l|ccccc|}
    \hline
    Steps & 1 & 2 & 4 & 7 & 10 \\
    \hline
    \plotblue{RAND} & 34.6 & 36.2 & 42.3 & 48.0 & 51.1 \\
    \plotorange{Ours} & 41.7 & 44.1 & 47.4 & 50.8 & 55.0 \\
    $\Delta$ & \textbf{\color{fgreen}+7.1} & \textbf{\color{fgreen}+7.9} & \textbf{\color{fgreen}+5.1} & \textbf{\color{fgreen}+2.8} & \textbf{\color{fgreen}+3.9} \\
    \hline
    \end{tabular}
    \caption{SPVCNN~\cite{eccv2020spvnas} on nuScenes~\cite{fong2022nuscenes}.}
    \label{tab:nuscenes}
\end{minipage}
\vspace{-25px}\end{table*}

\noindent \textbf{Disc Aggregation:} In Tab.~\ref{tab:disc_aggregation}, we compare our disc aggregation strategy of voxelwise pooling followed by discwise sum to 2D analogous methods of discwise mean, and discwise sum. We further compare the effects of different symmetric functions within our 2-stage setup.

Firstly we observe the limitations of common aggregation strategies. The sum performs similarly, if not better at low step counts to our proposed methods. As it is biased towards frames with high point counts, the high information yield at early frames allows easier learning of head classes. This trait later becomes its own demise as the distribution of labels becomes increasingly long tailed, limiting the models capability to learn difficult examples.

As expected, the max operation outperforms the min during the initial iterations as high density discs often lead to multi-class voxels that show a large variation of uncertainty. While again the high information yield at early frames helps, it comes with a detriment at later stages. As the min operation aims to maximize the minimum information, it expects each voxel to contribute substantially to the learning process, i.e. it favors purely uncertain voxels within the selected discs. These result in a more even distribution thus allowing the strategy to outperform others at later steps.



\noindent \textbf{Disc Selection:} In Tab.~\ref{tab:disc_selection}, we compare our disc sampling strategy of mixed-integer linear optimization to maximize total MI to (i) random selection of discs, and to (ii) the heuristic approach of highest point count sampling (HPCS), where each iteration we sample $N$ discs that are associated with the frames that contain the highest number of points. In all baselines, we prohibit intersections between discs to prevent trivial and harming solutions of selecting neighboring discs around the argmax~\footnote{
    Given a LiDAR sensor that operates at $10Hz$ and an ego-vehicle that moves in a straight line at $50km/h$, each disc will be separated by only $2m$. This is further reduces at lower velocities and while turning. Thus when intersections are allowed, if a disc at $x_i$ yields the highest uncertainty, it is very likely that the neighboring discs at $x_{i-1}$ or  $x_{i+1}$ would yield the second highest, resulting in a severe overlap of voxels.
}. As seen, our AL pipeline outperforms baseline methods with considerable margins.

In Tab.~\ref{tab:results} we also provide a classwise breakdown of the IoU for the final step. Here it can be seen that through intelligent disc selection, our AL approach allows substantial improvements for mainly tail classes such as bicycle, motorcycle, other-vehicle, person and bicyclist. In Fig.~\ref{fig:results} we also provide visual results that demonstrate this effect.

\noindent \textbf{Semi-Supervised Training:} In Tab.~\ref{tab:results} we further compare our active learning model trained on labeled points to our semi-supervised approach. As seen (in step 10) the performance is improved significantly ($1.7\%$ mIoU), with major improvements seen for classes such as bicycle, truck, person and the difficult other-ground category. Our AL method performs $99.8\%$ relative to the baseline model trained on a fully labeled dataset, with only 51 discs labeled ($0.5\%$ annotation cost of framewise labeling the entire dataset).


\noindent \textbf{Model Independence:} Our proposed discwise AL pipeline is completely model independent. In Tab.~\ref{tab:minkowski} we provide additional results to demonstrate this by showing that our approach continues to significantly outperform baseline random sampling while using MinkowskiNet~\cite{cvpr2019minkowski}.



\noindent \textbf{Other Datasets:} Our method works well against the random baseline for both ScribbleKITTI~\cite{Unal_2022_CVPR} and nuScenes~\cite{fong2022nuscenes} which we showcase in Tab.~\ref{tab:scribblekitti} and Tab.~\ref{tab:nuscenes} respectively.

\section{Conclusion}

In this work we tackle the problem of active learning for LiDAR semantic segmentation while considering common labeling techniques that minimize labeling times. To this end, we propose querying discs to utilize the sequential nature of LiDAR frames. With discwise AL, we label an accumulated region only once to have it labeled on all frames. We further devise a new acquisition function that tackles bias introduced due to the variability of local and discwise point counts. Later we construct a mixed-integer linear program to provide a general solution to disc selection that considers intersections. Finally we improve the performance of our model via a semi-supervised approach that utilizes a mean teacher. Through extensive studies we show that while simple, our discwise AL framework is highly effective and can boast high performance gains compared to baseline approaches.

\noindent \textbf{Limitations:} To enable sequential labeling objects must remain static. In cases where multiple varying class object paths collide in global coordinates, the joint object cloud must be labeled on a per frame basis. Time savings are still substantial as most of the environment is dominated by static structures. Still, in our proposed active learning framework, we do not consider the additional cost of labeling colliding moving objects. We hope future work can tackle this issue.

{\small
\bibliographystyle{IEEEtran}
\bibliography{egbib}
}

\end{document}